\pdfoutput=1
\documentclass[11pt]{article}

\usepackage[utf8]{inputenc}
\usepackage{lmodern}
\usepackage[T1]{fontenc}
\IfFileExists{microtype.sty}{\usepackage{microtype}}{}
\usepackage[letterpaper,margin=1in]{geometry}
\usepackage{amsmath,amssymb,bm}
\usepackage{graphicx}
\usepackage{booktabs}
\usepackage{tabularx}
\usepackage{array}
\usepackage{caption}
\usepackage{subcaption}
\usepackage{enumitem}
\usepackage{xcolor}
\usepackage[round,sort&compress]{natbib}
\usepackage[colorlinks=true,linkcolor=blue!50!black,citecolor=blue!50!black,urlcolor=blue!50!black]{hyperref}
\hypersetup{%
  pdftitle={Penumbra: Physics-Constrained Retrieval of XCO2 from OCO-2 Spectra Beyond the Cloud Screen},%
  pdfauthor={Antong Zhang, Jiani Yang, Vijay Natraj, Robert Nelson, Jiachen Yao, Xi Deng, Zhen Zhang, Elham Kianiharchegani, Sihe Chen, Anima Anandkumar, George Karniadakis}%
}

\graphicspath{{figures/}}

\newcommand{\cotwo}{CO$_2$}
\newcommand{\otwo}{O$_2$}
\newcommand{\hto}{H$_2$O}
\newcommand{\xco}{XCO$_2$}
\newcommand{\ppm}{\,ppm}
\newcommand{\lnxc}{\ln(y/c)}
\newcommand{\um}{\,\ensuremath{\mu}m}
\newcommand{\ci}[2]{\ensuremath{[#1,\,#2]}}  
\newcommand{\vect}[1]{\bm{#1}}
\newcolumntype{Y}{>{\raggedright\arraybackslash}X}
\newcolumntype{C}{>{\centering\arraybackslash}X}

\title{Penumbra: Physics-Constrained Retrieval of \xco{} from OCO-2 Spectra Beyond the Cloud Screen}

\newcommand{\aff}[1]{\textsuperscript{#1}}
\newcommand{\paperauthors}{%
  Antong Zhang\aff{1,2,*}, Jiani Yang\aff{2}, Vijay Natraj\aff{3}, Robert Nelson\aff{3},
  Jiachen Yao\aff{2}, Xi Deng\aff{2},\\
  Zhen Zhang\aff{1}, Elham Kianiharchegani\aff{1}, Sihe Chen\aff{2}, Anima Anandkumar\aff{2},
  George Karniadakis\aff{1}}
\newcommand{\paperaffiliations}{%
  \aff{1}Brown University\qquad
  \aff{2}California Institute of Technology\qquad
  \aff{3}Jet Propulsion Laboratory\\[0.25em]
  \aff{*}Corresponding author, antong\_zhang@brown.edu}
\makeatletter
\renewcommand{\maketitle}{%
  \begin{center}
    {\LARGE\bfseries\@title\par}
    \vspace{1.3em}
    {\normalsize\paperauthors\par}
    \vspace{0.7em}
    {\small\itshape\paperaffiliations\par}
  \end{center}
  \vspace{0.8em}}
\makeatother

\begin{document}
\maketitle

\begin{abstract}
\noindent
Satellite measurements of column-averaged carbon dioxide (\xco{}) constrain carbon sources and
sinks. However, the operational retrieval for NASA's OCO-2 is slow and rejects most soundings as
cloudy, leaving only about 10\,\% of land soundings over the Americas with an accepted value. Here
we present Penumbra, a physics-constrained neural retrieval that processes accepted,
quality-flagged, and cloud-screened soundings with a single network. Penumbra estimates \xco{}
jointly with a distribution of light paths that a fixed physical layer uses to reconstruct the
spectrum. We train the network to learn the spectral signature of \cotwo{} from simulations and the
behavior of real clouds from unlabeled observations, constraining neighboring soundings to agree on
\cotwo{} while preserving sensitivity to an injected \cotwo{} perturbation. On an independent test
year (2021) held out during development, Penumbra matches the bias-corrected operational product on
accepted soundings at ground stations (0.92 versus 0.91\ppm{} per overpass) without labeled real
spectra in network training. For flagged soundings and 100\,km averages of screened-out soundings
that retain \cotwo{} information, Penumbra reduces error by 12 to 27\,\% relative to a prior map,
whereas forcing the operational algorithm on those screened-out soundings fails to improve on that
map. Together, these recoveries expand usable coverage from 10 to 30\,\% of soundings while running
roughly two orders of magnitude faster per sounding. Finally, we show that the spectral difference
between screened-out soundings and their accepted neighbors is a measurable light-path effect that a
vertically resolved path model reproduces, unlike one-dimensional cloud simulations.
\end{abstract}

\section{Introduction}
\label{sec:intro}

Since 2014, the Orbiting Carbon Observatory-2 (OCO-2) has recorded near-infrared spectra of
reflected sunlight at a rate of almost one million soundings per day
\citep{crisp2017onorbit,eldering2017first} to infer the column-averaged dry-air mole fraction of
carbon dioxide, \xco{}. Because regional carbon sources and sinks impart column-averaged signatures
on the order of 1\ppm{} against a $\sim$410\ppm{} background, atmospheric flux inversions require
sub-percent precision and minimal regional bias
\citep{miller2007precision,crowell2019carbon,byrne2023national}. The operational Atmospheric \cotwo{}
Observations from Space (ACOS) algorithm attains this accuracy via optimal estimation with a
full-physics forward model \citep{odell2012acos,odell2018improved}. However, this framework carries
two major limitations. First, each retrieval requires several minutes on a single CPU core
\citep{taylor2023evaluating}; second, because the algorithm remains accurate only when scattering is
weak, cloud screens and a quality filter discard most observations
\citep{taylor2016orbiting,odell2018improved}. Over land in the Americas between 2017 and 2020, only
10.3\,\% of the 86.6~million soundings yielded an accepted value. These losses are also spatially
uneven: frequently cloudy regions, such as the Amazon, are preferentially excluded
\citep{odell2018improved}, which can bias surface fluxes inferred from the data
\citep{corbin2009assessing,liu2026impact}.

Existing learned retrievals have either relied on operational \xco{} labels or stopped at the cloud
screen (Section~\ref{sec:related}). While networks trained on operational products are fast, they
inherit both the coverage limits and the biases of their labels \citep{san2026emulating}.
Similarly, a network trained on atmospheric model fields was found to predict its targets largely
from location and season \citep{breon2022potential}. Approaches trained on synthetic spectra with
known \cotwo{}, in turn, have been evaluated only on simulated data
\citep{callesaldarriaga2026amortized} or scored against the operational product
\citep{xie2024fast}.

In this work, we introduce Penumbra, a physics-constrained retrieval that estimates \xco{} across
accepted, flagged, and screened-out soundings using a single neural network trained without
operational \xco{} labels. Our design starts from a basic radiative-transfer constraint: in cloudy
scenes, \cotwo{} abundance and photon path length are entangled in the spectrum, meaning that an
individual cloudy sounding determines \xco{} only when the relation between photon path lengths
across the \otwo{} and \cotwo{} bands is known. Rather than prescribing this cross-band relation
from an idealized cloud model, we learn it directly from observations. Specifically, simulations
with known \cotwo{} teach the network the spectral signature of gas abundance, while unlabeled real
spectra, subject to physical consistency constraints, teach it how real clouds alter light paths.

We evaluate three main claims on held-out observations. (1)~\emph{Accuracy without labeled training
spectra.} On accepted soundings at TCCON ground stations in the held-out 2021 test year, Penumbra
matches the bias-corrected operational product, even though real spectra enter network training
unlabeled. (2)~\emph{Coverage beyond the cloud screen.} Penumbra retrieves both flagged soundings
and, as 100\,km averages, a pre-defined pool of cloud-screened soundings that retain \cotwo{}
information; across all six 2021 evaluations, these estimates reduce error by 12 to 27\,\% compared
with a prior map, expanding the retrieved share from 10 to 30\,\% of soundings.
(3)~\emph{Computational speed.} By transferring clear-sky references across neighboring soundings,
Penumbra runs roughly two orders of magnitude faster per sounding than the operational retrieval. We
also trace the physical basis of these gains by directly measuring the spectral difference between
screened-out spectra and their accepted neighbors as a light-path effect
(Section~\ref{sec:disc-path}). Our study covers land soundings over the Americas, and we detail the
limits of each claim in Section~\ref{sec:limits}.

\section{Related Work}
\label{sec:related}

\paragraph{Operational retrieval and light-path methods.}
Operational OCO-2 \xco{} is retrieved with the ACOS optimal-estimation algorithm
\citep{odell2012acos,odell2018improved}, preceded by cloud screens \citep{taylor2016orbiting} and
followed by a quality filter and an empirical bias correction fitted to truth proxies, including
TCCON \citep{taylor2026oco}. On overpasses with at least 100 accepted soundings, version~11.1 land
retrievals agree with TCCON to a standard deviation of 0.85\ppm{} \citep{das2025comparisons}.
Recent studies quantify and correct residual 0.1 to 0.4\ppm{} 3-D cloud biases in accepted soundings
\citep{massie2021analysis,massie2023insights,mauceri2023correcting,chen2025mitigation}, or expand
usable retrievals by 14 to 21\,\% through refined bias corrections and quality filters
\citep{keely2023nonlinear,keely2025uncertainty}. However, these post-processing methods operate
only on converged retrievals and do not recover soundings discarded by the cloud screens. Rather
than modeling scatterers explicitly, another line of work parameterizes their net effect on photon
paths, as in the photon path-length probability density function (PPDF) method
\citep{bril2007parameterization,oshchepkov2008ppdf,oshchepkov2012effects,oshchepkov2013simultaneous}
and FOCAL \citep{reuter2017radiative,reuter2017application}. Our light-path layer belongs to this
family, but differs in two ways: it resolves the pressure level at which photons backscatter, and we
use it as a training constraint and scene diagnostic rather than as the direct estimator of \xco{}.

\paragraph{Learned retrievals.}
Neural networks trained on OCO-2 spectra using atmospheric model analyses as labels can match
operational precision on clear nadir soundings \citep{david2021xco2}. Yet subsequent analysis
showed that such models largely mimic their training distribution by inferring latitude and date
from the spectra, missing a localized plume captured by the full-physics retrieval
\citep{breon2022potential}. Operational emulators reproduce ACOS values within 0.9 to 1.4\ppm{} in
about 0.1\,ms per sounding \citep{san2026emulating}, but by construction they inherit the
operational product's coverage gaps and biases. Pre-training on ACOS outputs and fine-tuning on
TCCON lowers single-sounding error by 10\,\% below the operational product in a held-out year
\citep{keely2025conditional}; in a
related direction, \citet{feng2026selective} extend predictions to soundings lacking an operational
value, reporting errors of 1.9 to 2.5\ppm{} at withheld stations. Simulation-trained retrievals, by
contrast, have been validated either against operational outputs \citep[1.8\ppm;][]{xie2024fast} or
solely on synthetic spectra \citep{callesaldarriaga2026amortized}. Rather than adopting generic
domain-adaptation remedies from simulation-based inference
\citep{cranmer2020frontier,ward2022robust,wehenkel2025addressing}, we reconcile simulations with
real observations through two measurement-specific constraints: a consistency loss across unlabeled
spectra \citep{laine2017temporal} and a derivative constraint along the forward-model \cotwo{}
Jacobian \citep{simard1991tangent,czarnecki2017sobolev}.

\section{Data}
\label{sec:data}

\subsection{OCO-2 spectra and sounding classes}
\label{sec:classes}

We use version~11r Level~1B calibrated OCO-2 radiances \citep{oco2team2019l1b} for all land
soundings in North and South America from 2017 to 2021 across nadir, glint, and target viewing
modes (sounding counts by year and class are given in Appendix~\ref{app:tests},
Table~\ref{tab:data}). Each sounding comprises three spectral bands---the \otwo{} A band near
0.76\um{} and the weak and strong \cotwo{} bands near 1.61 and 2.06\um{}---paired with solar and
viewing geometry, meteorological analysis, and the operational prior \cotwo{} profile. Because this
prior already captures the secular trend and seasonal cycle of \cotwo{}, a retrieval is judged by
how accurately it recovers departures from the prior, which are typically on the order of 1\ppm.

Before running the full-physics inversion, the operational pipeline screens for clouds using two
fast preprocessors: the A-Band Preprocessor (ABP), which compares surface pressure retrieved from
the \otwo{} A band against the meteorological analysis, and the IMAP-DOAS Preprocessor (IDP), which
compares \cotwo{} and \hto{} columns retrieved independently from the weak and strong \cotwo{}
bands \citep{taylor2016orbiting}. Soundings passing both screens enter the full-physics retrieval,
where converged solutions receive a quality flag \citep{odell2018improved}. Accordingly, we
partition the 2017--2020 archive into \emph{accepted} soundings with a good-quality operational
value (quality flag~0; 10.3\,\%), \emph{flagged} soundings with a bad-quality operational value
(quality flag~1; 7.7\,\%), and \emph{unretrieved} soundings lacking a converged operational Lite
value (82.0\,\%; Fig.~\ref{fig:overview}a), most of which are rejected by the cloud screens.

Although 82.0\,\% of soundings lack a converged operational retrieval, a substantial subset of those
screened-out observations still retains meaningful spectral information on \cotwo{}. We define this
\emph{recoverable pool} using fixed preprocessor bounds: an IDP \cotwo{} band ratio between 0.98
and 1.06, an \hto{} band ratio between 0.7 and 1.2, and either an ABP clear-sky flag or an absolute
A-band surface-pressure departure below 50\,hPa (Appendix~\ref{app:tests}). Accounting for 12.4\,\%
of the archive, this pool exceeds the accepted class in size. Because this definition relies solely
on operational preprocessor outputs, the pool is fixed independently of our retrieval model.
Together, accepted, flagged, and pool soundings represent 30.4\,\% of the archive, three times the
accepted share; we determine the remaining 69.6\,\% to be too optically thick to recover.

For both aggregation and evaluation, we divide each orbit into consecutive 100\,km along-track
segments, termed \emph{patches}. Within the recoverable pool, 43\,\% of soundings lie in patches
containing at least five accepted soundings, while another 16\,\% have at least five accepted
soundings 50 to 200\,km away along the same orbit; the remaining 41\,\% are \emph{isolated}.

\subsection{Simulation}
\label{sec:simulation}

Our network learns the spectral imprint of \cotwo{} from simulated soundings with known \xco{}.
Each simulation inherits solar and viewing geometry, meteorological profiles, the prior \cotwo{}
profile, and the instrument model from a real \emph{anchor} sounding. We draw the \xco{} offset from
the prior uniformly within $\pm 12$\ppm{}, independently of all other parameters, while sampling
cloud, aerosol, and surface properties partly from broad distributions and partly guided by the
anchor's measured spectrum (Appendix~\ref{app:simulation}). Using the operational full-physics forward
model, we generate 278{,}164 simulated spectra on accepted, flagged, and pool anchors from
2017--2019, holding out 33{,}088 for validation. All simulated columns are plane-parallel, with
partial cloud cover represented as a linear mixture of clear and cloudy columns (the independent
pixel approximation) without horizontal photon transport, a simplification to which we return in
Section~\ref{sec:disc-path}.

\subsection{Reference data}
\label{sec:reference}

For independent ground truth, we use the Total Carbon Column Observing Network (TCCON), whose
direct-sun ground-based spectrometers are nearly immune to atmospheric scattering
\citep{wunch2011total,laughner2024total}. Of the seven stations operating in our study region during
2017--2021, five (all between 34$^\circ$N and 55$^\circ$N) have overpasses in 2021. An
\emph{overpass} pairs the soundings of one orbit inside a coincidence box around a station with
TCCON spectra recorded within one hour, adjusted to the prior profile and averaging kernel of each
sounding \citep{wunch2011method} (Appendix~\ref{app:tests}).

Because TCCON stations in our domain are sparse and confined to North America during 2017--2021, we
evaluate flagged and pool soundings elsewhere against nearby accepted operational retrievals. A
patch containing at least five accepted soundings whose operational values have a standard deviation
of at most 2.5\ppm{} receives a \emph{patch label}, defined as the mean of those values. Patches
with fewer accepted soundings instead receive an \emph{along-track label}, the mean of at least five
accepted soundings on the same orbit 50 to 200\,km away. Although both labels derive from
operational retrievals, they serve as reference truth strictly for soundings that the operational
product omits or flags; accepted soundings are evaluated against TCCON alone.

\section{Method}
\label{sec:method}

\subsection{Problem and overview}
\label{sec:problem}

\begin{figure}[!tbp]
\centering
\includegraphics[width=\textwidth]{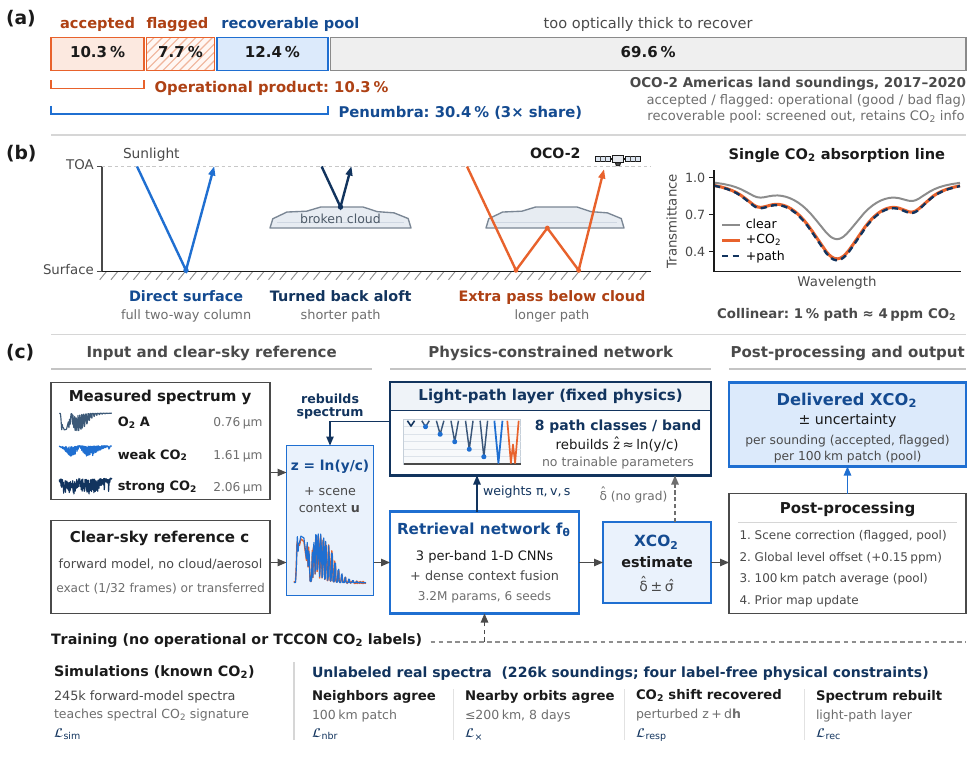}
\caption{Overview. (a) Sounding classes over land in the Americas, 2017--2020, with their shares
of the archive. (b) Three classes of light path under broken cloud. Path length and \cotwo{} deepen
the same absorption lines, which is why clouds bias the retrieval. (c) The Penumbra pipeline. The
measured spectrum $y$ is divided by a clear-sky reference $c$. A network estimates \xco{}, its
uncertainty, and the weights of a fixed light-path layer that reconstructs the spectrum. Training
combines simulations with known \cotwo{} and unlabeled real spectra under four physical constraints,
and four post-processing steps yield the delivered value.}
\label{fig:overview}
\end{figure}

In a clear atmosphere, sunlight traverses the atmospheric column twice, and absorption-line depths
measure the column of each gas along that geometric path: the \otwo{} A band yields the dry-air
column, the \cotwo{} bands yield the \cotwo{} column, and their ratio gives \xco{}. A cloud or
aerosol layer, however, alters the photon path (Fig.~\ref{fig:overview}b). Light backscattered
above the surface crosses only part of the column, whereas multiply scattered light traverses part
of it more than once. Monochromatic radiance in band $b$ is therefore an average over the photon
paths reaching the detector \citep{bril2007parameterization,funk2003photon},
\begin{equation}
  y(\lambda) \;=\; I_b(\lambda)\int p_b(\ell)\,
  \exp\!\bigl[-\tau(\lambda;\ell)\bigr]\,\mathrm{d}\ell ,
  \label{eq:paths}
\end{equation}
where $I_b$ collects the solar irradiance, surface reflectance, and non-absorbing scattering
continuum, $p_b$ is the distribution of paths $\ell$ in band $b$, and $\tau(\lambda;\ell)$ is the
gas optical depth along $\ell$. Within a \cotwo{} band, scaling either the \cotwo{} column or the
path length multiplies the same optical depth, making their spectral signatures nearly collinear:
increasing the path by one percent deepens the \cotwo{} lines roughly as much as adding 4\ppm{} of
\cotwo{}. Although the A band constrains path changes at 0.76\um, a single spectrum does not
uniquely determine how scattering modifies the path differently across the three bands because
cloud optical properties and surface reflectance vary strongly with wavelength.

This entanglement shapes the design of Penumbra (Fig.~\ref{fig:overview}c): we first remove from
each spectrum the clear-sky signal predicted by the forward model
(Section~\ref{sec:reference-transfer}), then train a network to estimate \xco{} alongside an
explicit light-path description while learning cross-band cloud behavior from unlabeled
observations (Sections~\ref{sec:network} and \ref{sec:labelfree}). Finally, we deliver per-sounding
estimates where individual spectra suffice and patch averages where spatially correlated cloud
errors dominate (Section~\ref{sec:product}).

\subsection{Clear-sky reference and spectral features}
\label{sec:reference-transfer}

For each sounding $i$, we compute the clear-sky spectrum expected with \cotwo{} at its prior
profile over a fixed Lambertian surface (albedo 0.2, without cloud or aerosol),
$c_i = F(\vect{g}_i,\vect{w}_i,\vect{a}_i)$, where $F$ is the operational forward model
\citep{odell2018improved}, $\vect{g}_i$ the solar and viewing geometry, $\vect{w}_i$ the
meteorological state, and $\vect{a}_i$ the prior \cotwo{} profile with column average $x^{a}_i$.
Computing this reference requires only a single forward pass without Jacobians.

The network takes as input the logarithmic ratio $z_i=\ln(y_i/c_i)$, which largely cancels factors
shared by the measurement and reference (solar Fraunhofer lines, instrument response, and prior
clear-sky gas absorption) while converting smooth multiplicative perturbations into additive terms.
From this ratio, we estimate the offset $\delta_i$ of \xco{} from $x^{a}_i$. Because we apply the
same division to real and simulated spectra, it also suppresses systematic differences between the
two domains. Specifically, we interpolate $z$ onto a fixed grid of 2{,}298 points and subtract a
per-band quadratic continuum polynomial, appending its coefficients to a 30-element scene-context
vector $\vect{u}_i$ (Appendix~\ref{app:training}) that by design omits geographic coordinates,
date, prior \xco{}, and cloud-screen values.

Although the forward run for $c_i$ dominates Penumbra's computational cost, neighboring soundings
share nearly the same clear-sky reference. We therefore compute $c$ exactly for only one sounding
per 32-frame window on each of the eight across-track footprints, roughly once every 70\,km.
Intermediate soundings borrow the nearest preceding and following exact references on the same
footprint, adjusted for differences in geometry and atmospheric state by learned transfer models
(Appendix~\ref{app:training}); whenever a sounding's surface pressure differs from that of its
references by more than 50\,hPa, we instead run an exact forward calculation. Across the archive,
these rules assign an \emph{exact} reference to 11\,\% of accepted, flagged, and pool soundings and
a \emph{transferred} reference to the remainder.

\subsection{Retrieval network and light-path layer}
\label{sec:network}

Our network $f_\theta$ maps $(z_i,\vect{u}_i)$ to an estimate $\hat\delta_i$, a standard deviation
$\hat\sigma_i$, and a vector of light-path parameters. Each spectral band passes through a
dedicated tower of one-dimensional convolutions; the resulting three band summaries and an
embedding of the scene context are then combined in two dense layers from which the output heads
branch (3.2~million parameters; Appendix~\ref{app:training}). For inference, we average an ensemble
of six networks trained from independent random seeds.

To enforce radiative-transfer structure, we pair the network with a fixed, differentiable
light-path layer implementing Eq.~\eqref{eq:paths} with no trainable parameters. Rather than a
continuous path distribution $p_b$, this layer assigns weights $\pi_{bk}$ over $K=8$ discrete path
classes defined by the scattering altitude: one class traverses zero absorbing gas, five
backscatter at successive pressure levels within the atmosphere, and the final two reach the
surface, once directly and once with an additional two-way pass through the lower troposphere.
Letting $q_{kj}\in\{0,1,2\}$ denote the number of two-way traversals of atmospheric layer $j$ by
class $k$, we compute the class optical depth in band $b$ and the reconstructed spectrum as
\begin{align}
  \tau_{bk}(\lambda) &= M\,s\sum_{j} q_{kj}\Bigl[\gamma_b\,
      \tau^{\text{gas}}_{bj}(\lambda) + v\,\tau^{\text{wv}}_{bj}(\lambda)\Bigr], \\
  \hat z_b(\lambda) &= \ln\Bigl[\mathcal{S}\ast\!\sum_{k=1}^{K}\pi_{bk}\,e^{-\tau_{bk}}\Bigr](\lambda)
      \;-\; \ln\Bigl[\mathcal{S}\ast e^{-\tau^{c}_{b}}\Bigr](\lambda) \;+\; \beta_b(\lambda).
  \label{eq:layer}
\end{align}
Here $M$ is the two-way geometric air-mass factor, $s$ is a common path-length scale factor within
$\pm 5\,\%$, and $v$ is a water-vapor scale factor. The terms $\tau^{\text{gas}}_{bj}$ and
$\tau^{\text{wv}}_{bj}$ represent the vertical optical depths of layer $j$ for the band's primary
absorber and for water vapor at the prior state; $\gamma_b$ equals $1+\hat\delta/x^{a}$ in the two
\cotwo{} bands and 1 in the A band, $\tau^{c}_{b}$ is the corresponding two-way clear-sky slant
optical depth from the same tables, $\mathcal{S}\ast$ denotes convolution with the instrument line
shape, and $\beta_b$ collects a quadratic continuum and linearized wavelength shift fitted in
closed form. From the shared trunk, the light-path head predicts the 24 class weights together with
$v$ and $s$.

During training, the spectral misfit between $z$ and $\hat z$ acts as a reconstruction loss on
simulated and real spectra alike, constraining learned cloud representations with known gas
spectroscopy; at inference, the predicted path weights and misfit provide scene diagnostics for
post-processing (Section~\ref{sec:product}). However, this physical layer does not determine \xco{}
on its own: because the per-band path weights in Eq.~\eqref{eq:layer} are unconstrained across
bands, path-length adjustments can absorb \cotwo{} variations, and variants that solved for \xco{}
directly through the layer or backpropagated the reconstruction loss into $\hat\delta$ did not
improve cloudy-sky accuracy. We therefore supply $\hat\delta$ to Eq.~\eqref{eq:layer} with
gradients detached, so the reconstruction term regularizes the shared representation rather than
pulling $\hat\delta$ directly.

\subsection{Learning from real spectra without labels}
\label{sec:labelfree}

While simulated spectra teach the network how \cotwo{} alters radiance, they cannot teach physics
absent from the simulator. Indeed, a network trained on simulations alone estimates the recoverable
pool roughly 1\ppm{} lower than accepted soundings in the same patch. Real spectra contain this
missing cloud behavior, but lack labels that are both accurate and independent of the operational
product. We therefore incorporate real observations through label-free physical constraints,
minimizing
\begin{equation}
  \mathcal{L} \;=\; \mathcal{L}_{\text{sim}}
  + \mathcal{L}_{\text{nbr}}
  + \alpha_{\times}\,\mathcal{L}_{\times}
  + \mathcal{L}_{\text{resp}}
  + \alpha_{\text{rec}}\,\mathcal{L}_{\text{rec}},
  \label{eq:loss}
\end{equation}
with weights $\alpha_{\times}=0.3$ and $\alpha_{\text{rec}}=0.2$. On simulated spectra,
$\mathcal{L}_{\text{sim}}$ is the squared error of $\hat\delta$ relative to the drawn offset. The
next three terms act strictly on real spectra, whereas $\mathcal{L}_{\text{rec}}$ acts on both real
and simulated spectra.

\emph{Neighbor consistency.} Following the small-area approximation used in OCO-2 validation
\citep{worden2017evaluation,odell2018improved}, \xco{} varies weakly across a 100\,km along-track
patch in most scenes, so spread among estimates within a patch primarily reflects retrieval error.
Accordingly, $\mathcal{L}_{\text{nbr}}$ penalizes the within-patch variance of estimates separately
for accepted, flagged, and pool soundings, avoiding any direct pull of pool estimates toward
accepted ones. Together, the label-free terms nevertheless remove most of the offset between the
two classes (Section~\ref{sec:res-ablation}).

\emph{Cross-orbit consistency.} Because residual error on accepted soundings is largely spatially
correlated across an entire overpass, it remains invisible to within-patch variance. To expose it,
$\mathcal{L}_{\times}$ applies a Huber loss to the difference between accepted patch-mean
departures $\hat\delta$ from distinct orbits separated by at most 200\,km and 8~days.

\emph{\cotwo{} response constraint.} Consistency losses alone admit a trivial solution: an
estimator insensitive to \cotwo{} agrees closely with its neighbors. When we trained with accepted
neighbor consistency alone, the \cotwo{} response (Section~\ref{sec:setup}) collapsed to 0.06. We
prevent this collapse by anchoring the network's local sensitivity without prescribing its value.
Let $\vect{h}_i=\partial \ln c_i/\partial\text{\xco}$ denote the clear-sky spectral Jacobian with
respect to a uniform column \cotwo{} perturbation, obtained either from the forward model or from
an auxiliary predictor network (Appendix~\ref{app:training}). For a real spectrum $z_i$ and a
random-sign step $d$ between 0.5 and 3\ppm{}, we require the prediction on the perturbed spectrum
$z_i+d\,\vect{h}_i$ to shift by $d$:
\begin{equation}
  \mathcal{L}_{\text{resp}} \;=\;
  \biggl(\frac{f_\theta(z_i+d\,\vect{h}_i,\vect{u}'_i)-f_\theta(z_i,\vect{u}_i)}{d}-1\biggr)^{2},
\end{equation}
where $\vect{u}'_i$ is the scene context recomputed for the perturbed spectrum.

\emph{Reconstruction.} The final term, $\mathcal{L}_{\text{rec}}$, applies a robust penalty to the
noise-weighted misfit between $z$ and the spectrum reconstructed by the light-path layer, evaluated
on both real and simulated inputs.

For unlabeled real training data, we draw 226{,}437 accepted, flagged, and pool soundings from 2017
and 2018 across 18{,}614 patches, excluding all patches near TCCON stations. Neither operational
nor TCCON \xco{} values enter Eq.~\eqref{eq:loss}; operational products influence network training
solely through sounding classification and patch selection.

\subsection{Post-processing and product generation}
\label{sec:product}

Four post-processing steps convert raw network predictions into the delivered product
(Appendix~\ref{app:training}).

\emph{Scene-dependent correction.} In cloudy scenes, network estimates retain predictable
scene-dependent biases. For flagged and pool soundings, we model and subtract this bias using
gradient-boosted trees \citep{friedman2001greedy} driven by cloud-screen outputs, predicted
uncertainty, light-path layer outputs, linearized retrieval diagnostics, and scene and patch
descriptors. As training target, we use each sounding's estimate minus the mean estimate of
accepted soundings in the same patch, or minus the along-track label when the patch lacks accepted
soundings. Because soundings within a patch share essentially the same \xco{}, this model learns
scene-dependent errors rather than \xco{} itself.

\emph{Global bias offset.} We add a single global constant, $+0.15$\ppm, so that accepted overpass
means agree with TCCON on average across 2017--2020.

\emph{Aggregation.} Within each patch, we combine corrected pool estimates by inverse-variance
weighting and report patch averages for patches containing at least ten pool soundings. Because the
dominant residual error is spatially correlated across a cloudy scene, patch uncertainty does not
shrink as the inverse square root of sample count; we therefore predict a correlated scene-error
variance and scale it to match the correlated error observed at TCCON stations during 2017--2020.

\emph{Bayesian prior update.} Finally, we combine each prior-subtracted estimate $\hat\delta$ and
its estimated standard deviation $\sigma$ with a climatological prior map $\mu$ of standard
deviation $\sigma_\mu$ via
$\hat\delta^{\text{post}} = (\hat\delta/\sigma^2+\mu/\sigma_\mu^2)/(1/\sigma^2+1/\sigma_\mu^2)$,
applied per sounding for accepted and flagged observations and per patch for the pool. This prior
map captures the smooth departure of \xco{} from the ACOS prior as a function of latitude,
longitude, and month, with a linear temporal trend fitted to accepted operational values from
2017--2020. Because combining with a prior shrinks estimates toward the map and reduces measurement
sensitivity, we evaluate performance both before this update (unshrunk single-sounding estimates)
and after it (the delivered product).

In summary, the neural network itself trains without operational or TCCON \xco{} labels; from the
operational system it uses only the forward model, sounding classifications, training-patch
selection, and scene statistics for simulations. Post-processing uses operational products more
broadly, drawing on preprocessor outputs and neighboring retrievals as predictors, accepted
operational values for the prior map and along-track labels, and TCCON measurements for the global
bias offset and uncertainty scaling.

\section{Experimental Setup}
\label{sec:setup}

\paragraph{Splits and evaluation protocol.}
\label{sec:protocol}
None of the 2021 observations influenced model design or tuning. Simulations use 2017--2019
anchors, label-free training uses real spectra from 2017 and 2018, and all post-processing models
are fitted on 2017--2020 using out-of-year estimates from network variants trained without the
target year (Appendix~\ref{app:tests}). This leave-one-year-out scheme enables evaluation across
2017--2020, which we designate the \emph{development years} because they guided design choices. By
contrast, the 2021 archive was held out as an untouched (\emph{sealed}) test year throughout
development: we finalized the network, post-processing pipeline, baselines, and evaluation sampling
rules before scoring 2021 once (Appendix~\ref{app:tests}).

\paragraph{Truth and metrics.}
We evaluate the three sounding classes separately, scoring overpass means against TCCON near ground
stations and patch values elsewhere (against the patch label when at least five accepted soundings
are present, and against the along-track label otherwise). An overpass or patch qualifies in a
given class if it contains at least 5 accepted, 3 flagged, or 10 pool soundings, with estimates and
reference truths expressed as departures from each sounding's ACOS prior. Performance of the
delivered product is summarized by root-mean-square error (RMSE). When comparing retrievals
directly, however, RMSE alone can mislead: an estimator that under-responds to real \cotwo{}
variations stays closer to the prior and exhibits artificially low scatter. We therefore also report
the \emph{error after rescaling}, $\varepsilon = \sigma_t\sqrt{1-\rho^2}$, where $\sigma_t$ is the
standard deviation of the truth and $\rho$ its correlation with the estimate; this quantity is the
residual standard deviation from a linear regression of the truth onto the estimate, making it
invariant to any constant offset or multiplicative scaling. To probe physical sensitivity directly,
the \emph{\cotwo{} response} adds the clear-sky spectral signature of a 1\ppm{} \xco{} perturbation
to a real spectrum and records the fraction recovered by the network.

\paragraph{Baselines.}
The \emph{operational product} refers to bias-corrected \xco{} from the OCO-2 version~11.2r Lite
files \citep{oco2team2024lite}. Two spectrum-free baselines establish no-measurement benchmarks:
the operational \emph{ACOS prior} of each sounding and the climatological \emph{prior map} from
Section~\ref{sec:product}. Because this map represents what can be inferred without inspecting the
spectrum, flagged and pool retrievals add value only where they outperform it. For pool soundings,
the operational full-physics algorithm is also run directly as a \emph{forced operational retrieval}.
Two \emph{label-supervised} baselines sharing Penumbra's architecture and inputs are trained on
operational \xco{} from accepted soundings, with one subsequently fine-tuned on TCCON following the
protocol of \citet{keely2025conditional}. A simplified \emph{per-sounding Bayesian fit} with a
surrogate forward operator, modeled after \citet{meng2022learning}, is evaluated at ground stations
(Appendix~\ref{app:training}).

\paragraph{Fair comparison and statistics.}
Comparisons between delivered Penumbra estimates and another method either apply the same prior
update to both sides (after scaling each method's reported uncertainty identically against
development-year TCCON errors) or evaluate both methods prior to that step. Confidence intervals at
the 95\,\% level come from 2{,}000 paired bootstrap resamples over overpasses or patches. Retraining
the ensemble with different random seeds shifts these errors by roughly 0.01\ppm{} on patches and
0.02\ppm{} at stations; smaller differences are practically indistinguishable.

\section{Results}
\label{sec:results}

Table~\ref{tab:main} summarizes the headline results on held-out 2021 observations across our three
claims.

\begin{table}[!tbp]
\centering
\caption{Main results (ppm). Differences are Penumbra minus the comparator with 95\,\% bootstrap
intervals. (A) Overpass means of accepted soundings against TCCON at five stations, where each
product receives one global offset and RMSE is computed after the same prior update. (B) RMSE of
the delivered values of flagged and pool soundings against the truth named in the second column.
(C) Equal-footing comparison of Penumbra's uncorrected network estimates (without scene-dependent
correction) against operational retrievals of the same soundings: RMSE after the same prior update
for flagged soundings, and error after rescaling for the forced operational retrieval on the pool.
Appendix~\ref{app:tables} gives the number of overpasses and patches behind every row.}
\label{tab:main}
\small
\setlength{\tabcolsep}{5pt}
\begin{tabularx}{\textwidth}{@{}l Y r r l@{}}
\toprule
\multicolumn{5}{@{}l}{\textbf{A. Accepted soundings at TCCON stations}}\\
\midrule
Years & Metric & Penumbra & Operational & Difference \\
\midrule
2021 (sealed) & error after rescaling & 0.92 & 0.91 & $+0.01$ \ci{-0.06}{+0.07} \\
 & RMSE & 0.83 & 0.85 & $-0.02$ \ci{-0.06}{+0.01} \\
 & error after rescaling, operational retrieval before its bias correction & 0.92 & 1.06 & $-0.15$ \ci{-0.23}{-0.06} \\
2017--2020 & error after rescaling & 0.78 & 0.86 & $-0.09$ \ci{-0.13}{-0.05} \\
\quad (development) & RMSE & 0.73 & 0.78 & $-0.05$ \ci{-0.08}{-0.02} \\
\midrule
\multicolumn{5}{@{}l}{\textbf{B. Flagged and pool soundings, 2021, against the prior map}}\\
\midrule
Soundings & Truth & Penumbra & Prior map & Difference \\
\midrule
Flagged & accepted soundings in the same patch & 0.79 & 1.08 & $-0.29$ \ci{-0.33}{-0.25} \\
Flagged & ground stations (TCCON) & 0.91 & 1.14 & $-0.23$ \ci{-0.34}{-0.12} \\
Pool & accepted soundings in the same patch & 0.83 & 1.10 & $-0.27$ \ci{-0.30}{-0.23} \\
Pool & ground stations (TCCON) & 1.03 & 1.17 & $-0.14$ \ci{-0.25}{-0.03} \\
Pool & accepted soundings 50--200\,km along the orbit & 1.04 & 1.19 & $-0.15$ \ci{-0.18}{-0.12} \\
 & \quad the same, no accepted sounding in the patch & 1.07 & 1.22 & $-0.15$ \ci{-0.19}{-0.11} \\
\midrule
\multicolumn{5}{@{}l}{\textbf{C. Against operational values of the same soundings, 2021}}\\
\midrule
Soundings & Truth and metric & Penumbra & Operational & Difference \\
\midrule
Flagged & ground stations (TCCON), RMSE & 0.95 & 1.12 & $-0.17$ \ci{-0.26}{-0.08} \\
Flagged & same patch, RMSE & 0.84 & 0.82 & $+0.02$ \ci{-0.01}{+0.05} \\
Pool & ground stations, forced operational retrieval, error after rescaling & 1.24 & 1.46 & $-0.22$ \ci{-0.32}{-0.13} \\
\bottomrule
\end{tabularx}
\end{table}

\subsection{Simulations}
\label{sec:res-sim}

Because a learned retrieval can achieve low error on real data simply by returning a smooth
function of location and season \citep{breon2022potential}, we first evaluate the network where the
true state is known. On the 33{,}088 held-out simulations, whose offsets from the prior are drawn
uniformly within $\pm 12$\ppm, the ensemble recovers the true offset with an RMSE of 2.41\ppm{}
overall (spanning cloud optical depths up to 100), 1.58\ppm{} for optically thin clouds
($\tau \le 0.3$), and 1.14\ppm{} for cloud-free scenes. Moreover, this sensitivity survives
training on real spectra: the median \cotwo{} response is 0.98 for a network trained on simulations
alone and 0.97 for Penumbra. In contrast, our two label-supervised baselines respond at 0.73 and,
after fine-tuning on TCCON, at just 0.37 (Table~\ref{tab:ablation}), indicating that they partly
rely on non-\cotwo{} scene and spectral cues correlated with their training labels rather than the
\cotwo{} absorption lines.

\subsection{Accepted soundings at ground stations}
\label{sec:res-accepted}

\begin{figure}[!tbp]
\centering
\includegraphics[width=\textwidth]{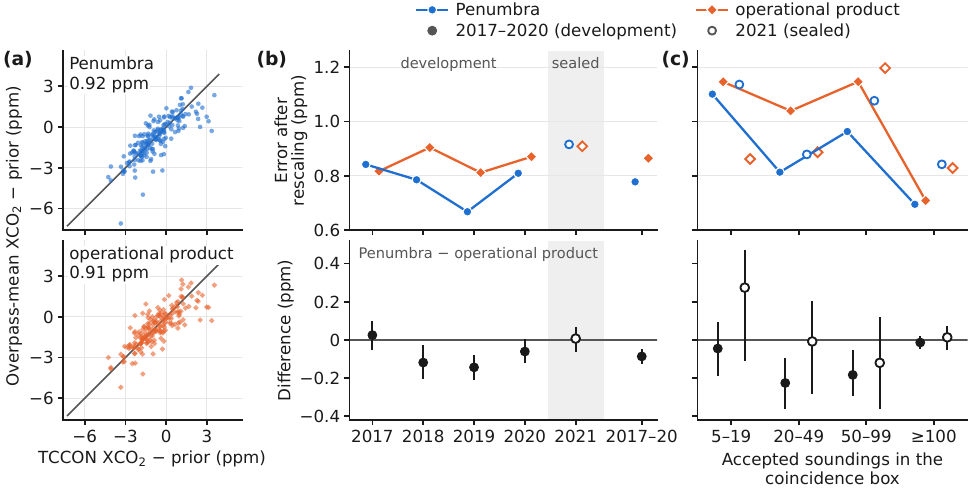}
\caption{Overpass means of accepted soundings at five TCCON stations. (a) Held-out year 2021,
plotting \xco{} minus the prior versus TCCON minus the prior, with the 1:1 line and the error after
rescaling. (b) Error after rescaling by year and pooled over the development years (top), together
with the difference (Penumbra minus operational product) and 95\,\% bootstrap intervals (bottom).
(c) Same comparison across four bins of the number of accepted soundings in the coincidence box for
the development years (filled) and 2021 (open).}
\label{fig:accuracy}
\end{figure}

In 2021, Penumbra achieves parity with the bias-corrected operational product on accepted
soundings. Across five TCCON stations, the error after rescaling of overpass means is 0.92\ppm{}
for Penumbra and 0.91\ppm{} for the operational product, a difference of $+0.01$\ppm{}
\ci{-0.06}{+0.07} (Fig.~\ref{fig:accuracy}a, Table~\ref{tab:main}A).

This parity is informative because the two estimates arrive by different routes. The operational
product owes part of its accuracy to an empirical bias correction fitted to TCCON and other truth
proxies \citep{odell2018improved,taylor2026oco}. Before that correction, the operational retrieval
has an error after rescaling of 1.06\ppm{} on the same overpasses, 0.15\ppm{} higher than Penumbra
\ci{0.06}{0.23}. By contrast, Penumbra's accepted-sounding estimates undergo no scene-dependent
correction, and error after rescaling is invariant to the single global offset, meaning this
0.92\ppm{} accuracy comes directly from the label-free network. Furthermore, the accepted class is
defined by the operational quality filter itself, so the comparison evaluates soundings selected
specifically for the operational product.

In the development years (2017--2020), where each year is evaluated by a network variant trained
without that year's spectra, Penumbra outperforms the operational product at the same stations
(0.78 versus 0.86\ppm, a difference of $-0.09$\ppm{} \ci{-0.13}{-0.05}; Fig.~\ref{fig:accuracy}b,
Table~\ref{tab:peryear}). Because real spectra are unlabeled during training, these years are free
of target leakage; nonetheless, because they guided model design, we base our primary claim on
2021. Stratifying overpasses by sample size locates the 2017--2020 advantage in sparser overpasses
containing 20 to 99 accepted soundings in the coincidence box, whereas the two methods tie on
well-sampled overpasses ($\ge 100$ accepted soundings; Fig.~\ref{fig:accuracy}c).

\subsection{Flagged and screened-out soundings}
\label{sec:res-coverage}

\begin{figure}[!t]
\centering
\includegraphics[width=\textwidth]{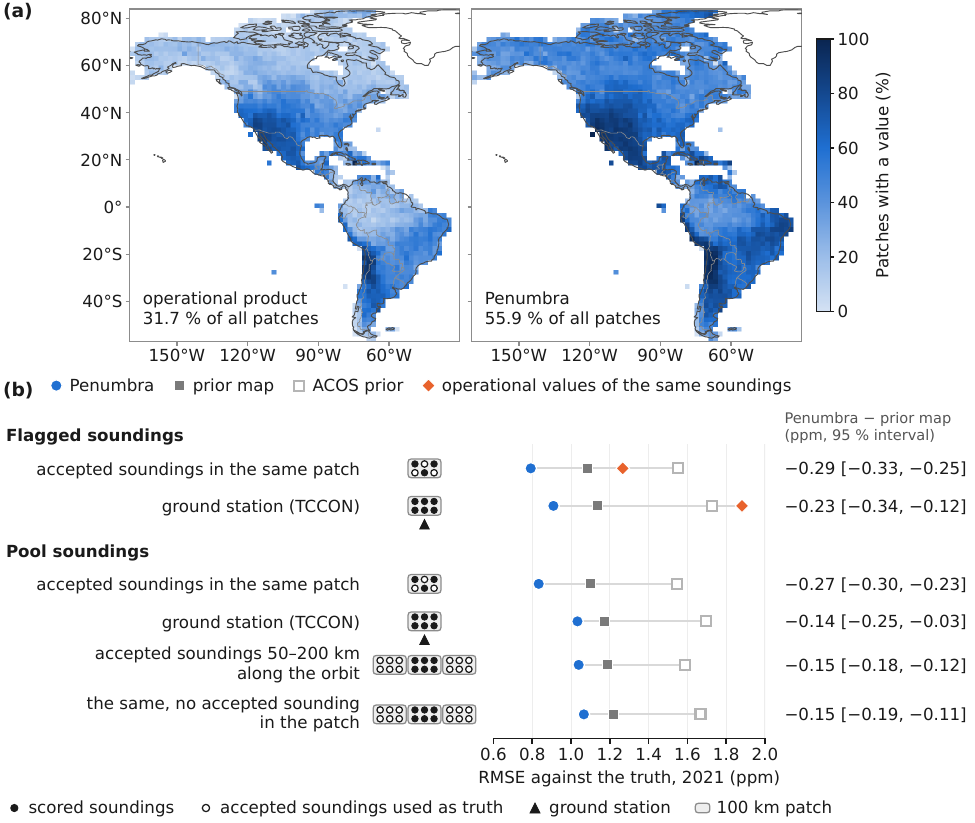}
\caption{Beyond the cloud screen. (a) Share of 100\,km patches with a value in 2017--2020
($2^\circ\times 2^\circ$ cells) for the operational product (patches with at least one accepted
sounding) and for Penumbra (patches with at least one accepted or flagged sounding, or at least ten
pool soundings). (b) RMSE of the delivered Penumbra values in 2021 for each class and reference
truth, alongside the prior map, the ACOS prior, and (for flagged soundings) operational values
before prior updating. Pictograms indicate the truth source for each test; numbers on the right
give Penumbra minus the prior map with 95\,\% bootstrap intervals.}
\label{fig:coverage}
\end{figure}

Outside the accepted class, the central question shifts to whether a spectrum-derived estimate
improves on what is known without the spectrum. In 2021, Penumbra outperforms the prior map across
all six rows of Table~\ref{tab:main}B (Fig.~\ref{fig:coverage}b).

\paragraph{Flagged soundings.}
Evaluated against the accepted soundings of the same patch, delivered flagged values achieve an
RMSE of 0.79\ppm, compared with 1.08\ppm{} for the prior map; at TCCON, the corresponding errors
are 0.91 and 1.14\ppm. We also compared Penumbra's uncorrected network estimates for flagged
soundings against operational quality-flag~1 retrievals on an equal footing
(Table~\ref{tab:main}C). At TCCON, Penumbra outperforms the operational retrieval both before
either side uses the prior map (1.09 versus 1.30\ppm{} in error after rescaling) and after both
undergo the same prior update (0.95 versus 1.12\ppm{} RMSE). On patch labels, which are themselves
constructed from operational values, the two perform comparably after the same prior update (0.84
versus 0.82\ppm{}, difference $+0.02$\ppm{} \ci{-0.01}{+0.05}); adding Penumbra's scene-dependent
correction reduces this error to 0.79\ppm.

\paragraph{The recoverable pool.}
Pool soundings lack converged operational retrievals altogether. In patches containing at least
five accepted soundings, the pool estimate reaches an RMSE of 0.83\ppm{} relative to the patch
label, compared with 1.10\ppm{} for the prior map. Evaluated at TCCON, which is independent of the
operational product, it achieves 1.03 versus 1.17\ppm. In patches with fewer than five accepted
soundings, the pool error is 1.04\ppm{} relative to the along-track label versus 1.19\ppm{} for the
map. The gain over the map is thus largest adjacent to accepted soundings ($-0.27$\ppm) and roughly
half as large where accepted soundings are scarce or absent ($-0.15$\ppm); this improvement holds
across all four regions and seasons (Table~\ref{tab:region}).

\paragraph{What the pool adds.}
A pool estimate could outperform the prior map yet remain redundant if nearby accepted soundings
already constrain the local \xco{}. To test this possibility, we withheld the accepted soundings
along a segment of orbit and predicted their mean from the prior map and the remaining accepted
soundings on that orbit. Including local pool values lowers the prediction RMSE by 0.09\ppm{}
across a 20\,km withheld segment and by 0.10 to 0.20\ppm{} for entire patches whose nearest
accepted soundings lie 100 to 800\,km away (Appendix~\ref{app:tables}). The operational algorithm,
by contrast, fails to recover this pool. When we ran the full-physics retrieval on the pool
soundings near TCCON stations, it converged on 60\,\% of them; on those converged soundings, its
raw (pre-bias-correction) overpass means yield an error after rescaling of 1.46\ppm. That error
exceeds both raw Penumbra estimates of the same soundings before post-processing (1.24\ppm;
Table~\ref{tab:main}C) and the prior map (1.14\ppm), whereas delivered Penumbra values improve on
the map (1.02\ppm).

\paragraph{Coverage and uncertainty.}
\label{sec:res-unc}
Combining accepted, flagged, and pool soundings, Penumbra utilizes 30\,\% of land soundings where
the operational product accepts 10\,\%. Because pool estimates are reported per patch,
Fig.~\ref{fig:coverage}a maps the fraction of patches with a value, which rises from 31.7 to
55.9\,\% overall, from 24.8 to 49.3\,\% over the Amazon, and from 13.5 to 47.0\,\% north of
60$^\circ$N. Most newly covered patches are isolated from accepted soundings, however; our
validation there remains indirect (Section~\ref{sec:limits}). Each delivered value also carries a
reported uncertainty; 62 to 76\,\% of reference truths fall within one standard deviation and 89 to
96\,\% within two (Table~\ref{tab:unc}).

\subsection{Computational speed}
\label{sec:res-speed}

\begin{figure}[!tbp]
\centering
\includegraphics[width=\textwidth]{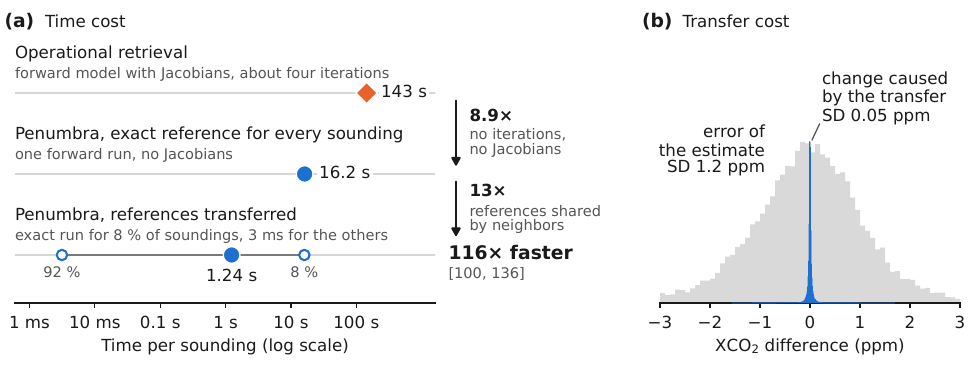}
\caption{Computational speed and reference-transfer accuracy. (a) Time cost: mean runtime per
sounding from a head-to-head benchmark on the same 2021 soundings and compute node. With
transferred references, the 1.24\,s mean combines an exact forward run for 8\,\% of soundings with
about 3\,ms for the rest (open circles); the interval in brackets gives the 95\,\% bootstrap
confidence interval on the speedup factor. (b) Transfer cost: shift in estimated \xco{} when
transferring rather than computing the reference exactly (blue), compared with single-sounding
accepted errors at TCCON (gray). Both histograms are normalized to equal peak height.}
\label{fig:speed}
\end{figure}

We benchmarked Penumbra directly against the operational retrieval on the same 2{,}014 soundings of
2021 on the same compute node (single CPU core for radiative transfer and reference transfer;
network inference adds 0.02\,ms on a GPU or 6\,ms on a single CPU core, leaving the mean
essentially unchanged; Fig.~\ref{fig:speed}a). Averaging 143\,s per sounding, the operational
retrieval spends most of its runtime executing the forward model with Jacobians roughly four times
before converging. Computing an exact clear-sky reference for each sounding lowers Penumbra's mean
runtime to 16.2\,s, dominated by the single forward pass. With transferred references, just 8\,\% of
soundings require a forward run while the remainder take a few milliseconds each, reducing the mean
time to 1.24\,s, a speedup of 116 times over the operational retrieval \ci{100}{136}. Across the
full archive, where 11\,\% of soundings require an exact run, the speedup is roughly 80-fold.
Reference transfer incurs negligible accuracy loss: across the 51{,}165 evaluation soundings using
a transferred reference, estimates shift with a standard deviation of 0.05\ppm{} and a mean of at
most 0.002\ppm, small relative to the retrieval error itself (Fig.~\ref{fig:speed}b). Because these
timings exclude the fast operational preprocessors (ABP and IDP), and our post-processing uses
operational sounding classifications and neighboring surface albedos, Penumbra as evaluated here
relies on upstream operational outputs.

\subsection{Ablations and baselines}
\label{sec:res-ablation}

\begin{table}[!tbp]
\centering
\caption{Ablations and baselines in 2021. RMSE (ppm) of overpass or patch values after the same
prior update (the Bayesian fit uses its own posterior), matching the evaluation populations of
Table~\ref{tab:main}. The \cotwo{} response is the median on development soundings, and the two
supervised rows are the label-supervised baselines. The bold bottom row adds the scene-dependent
correction and aggregation of Section~\ref{sec:product} and is not like-for-like with the rows
above.}
\label{tab:ablation}
\small
\setlength{\tabcolsep}{4pt}
\begin{tabularx}{\textwidth}{@{}Y r r r r r r r@{}}
\toprule
 & \multicolumn{3}{c}{Ground stations} & \multicolumn{3}{c}{Patches} & \\
\cmidrule(lr){2-4}\cmidrule(lr){5-7}
Method & Accepted & Flagged & Pool & Flagged & \shortstack[r]{Pool, same\\patch} &
 \shortstack[r]{Pool, along\\the orbit} & \shortstack[r]{\cotwo\\response} \\
\midrule
Penumbra (network estimate) & 0.83 & 0.95 & 1.07 & 0.84 & 0.92 & 1.08 & 0.97 \\
\quad without light-path layer & 0.84 & 0.97 & 1.10 & 0.85 & 0.91 & 1.07 & 0.97 \\
\quad simulations only & 0.87 & 1.04 & 1.11 & 0.89 & 1.01 & 1.12 & 0.98 \\
\quad simulations + cloud transplant & 0.88 & 1.05 & 1.12 & 0.86 & 0.97 & 1.10 & 0.99 \\
\midrule
Supervised, operational values & 0.85 & 1.07 & 1.32 & 0.80 & 1.08 & 1.26 & 0.73 \\
Supervised, tuned on TCCON & 0.84 & 0.99 & 1.10 & 0.93 & 1.02 & 1.12 & 0.37 \\
Per-sounding Bayesian fit & 1.14 & 1.89 & 3.66 & -- & -- & -- & -- \\
Operational product & 0.85 & -- & -- & -- & -- & -- & -- \\
Prior map alone & 1.04 & 1.14 & 1.17 & 1.08 & 1.10 & 1.19 & -- \\
\midrule
\textbf{Penumbra, delivered product} & \textbf{0.83} & \textbf{0.91} & \textbf{1.03} & \textbf{0.79} & \textbf{0.83} & \textbf{1.04} & -- \\
\bottomrule
\end{tabularx}
\end{table}

Table~\ref{tab:ablation} compares individual training components and baselines after applying the
same prior update, pointing to four conclusions. First, clear-sky reference division and simulated
supervision anchor performance on accepted soundings, whereas the label-free terms govern the
cloudy populations. A network trained purely on simulations is not significantly worse than the
full model on accepted overpasses (0.87 versus 0.83\ppm), yet it degrades markedly on flagged and
pool soundings. On development data, adding the label-free terms eliminates its roughly 1\ppm{}
offset between pool soundings and accepted neighbors; in 2021, those terms reduce error on pool
patches next to accepted soundings from 1.01 to 0.92\ppm. Reference division is equally important:
a label-supervised network reading spectra without it exhibits a \cotwo{} response of 0.24 rather
than 0.73.

Second, adding light-path reconstruction yields modest accuracy gains at ground stations (e.g.,
0.83 versus 0.84\ppm{} on accepted soundings and 0.95 versus 0.97\ppm{} on flagged soundings;
Table~\ref{tab:ablation}), while differences after full post-processing remain within seed-to-seed
variation except on flagged patches. The layer's primary value therefore lies in physical
interpretability (Section~\ref{sec:disc-path}) and the scene diagnostics it provides without extra
computation. Third, the cloud signature learned from real spectra can be isolated and transplanted:
adding the measured spectral difference between pool soundings and their accepted neighbors onto
cloud-free \emph{simulated} spectra captures part of the cloudy-sky improvement (lowering pool
patch RMSE from 1.01 to 0.97\ppm{} and flagged patch RMSE from 0.89 to 0.86\ppm), though direct
label-free training on real spectra remains stronger across all populations.

Fourth, real-label supervision helps where labels exist and degrades elsewhere. Our baseline
fine-tuned on TCCON matches Penumbra on accepted soundings at the stations used for tuning (0.84
versus 0.83\ppm), yet trails across all three patch populations away from stations. Similarly, the
network trained on operational values improves over Penumbra exclusively on flagged patches, whose
evaluation labels are themselves derived from operational retrievals. Finally, our simplified
per-sounding Bayesian fit lags on accepted soundings and fails on the pool, consistent with a
surrogate forward operator that inherits the limitations of simulated clouds.

\section{Discussion}
\label{sec:discussion}

\subsection{The cloud effect is a measurable light-path effect}
\label{sec:disc-path}

\begin{figure}[!tbp]
\centering
\includegraphics[width=0.86\textwidth]{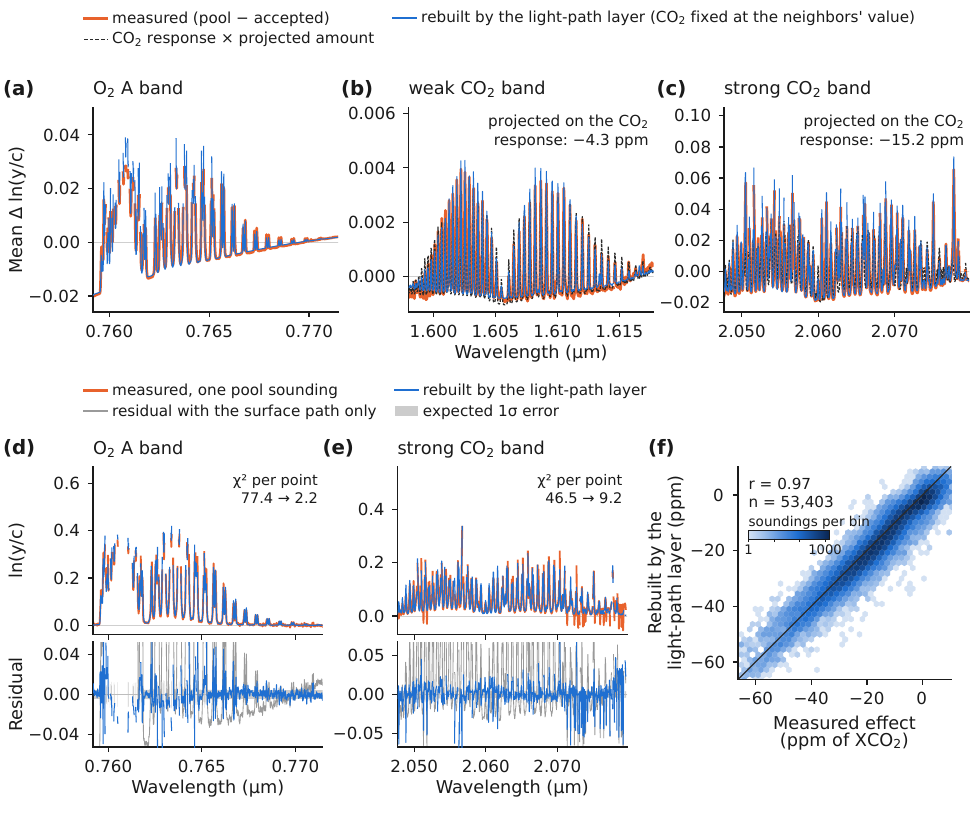}
\caption{The cloud effect is a light-path effect (development data, 2017--2018). (a--c) Mean
continuum-removed difference in $\lnxc$ between 53{,}403 pool soundings and accepted soundings
within the same 100\,km patch (orange), alongside the difference reconstructed by the light-path
layer with \cotwo{} fixed at the neighbors' value (blue). The dashed curve shows the spectral
response to \xco{}, scaled by projecting the measured difference onto it. (d, e) A pool sounding
with median misfit (Arizona, 24 December 2017), showing measured $\lnxc$, the reconstructed
spectrum, and the residual below; the gray curve shows the residual when restricting light to the
surface-reaching path. The weak \cotwo{} band, where the fit changes little, is omitted.
(f) Projection of the reconstructed difference onto the \xco{} response versus that of the measured
difference across all 53{,}403 soundings, with the 1:1 line.}
\label{fig:lightpath}
\end{figure}

Why can a neural network recover \xco{} from soundings discarded by the cloud screen, and why must
it see real spectra to do so? The answer appears directly in a model-free observable. Within a
100\,km patch, a pool sounding and its accepted neighbors sample nearly identical \cotwo{};
consequently, their spectral difference in $\lnxc$ isolates the imprint of cloud on the
measurement. We computed this difference across 53{,}403 pool soundings in 1{,}626 patches from
2017--2018 (Fig.~\ref{fig:lightpath}a--c). Projected onto the \xco{} spectral response, the mean
difference corresponds to $-14.5$\ppm, meaning that attributing the full signal to gas abundance
would bias the pool 14.5\ppm{} below its neighbors. Moreover, the deficit concentrates in the
strong \cotwo{} band ($-15.2$\ppm{} versus $-4.3$\ppm{} in the weak band) and grows with the cloud
screen's \cotwo{} band ratio. By comparison, differencing accepted soundings against their own
accepted neighbors yields $0.00 \pm 0.03$\ppm.

That clouds act through optical path length underlies both the cloud screens and the path-length
retrievals reviewed in Section~\ref{sec:related}. Here we quantify the magnitude of that effect
just beyond the screen and test a path parameterization capable of reproducing it. Fitting the
light-path layer to each pool spectrum with \cotwo{} held fixed at the neighbors' value reduces
median reduced $\chi^2$ from 7.0 to 2.1 in the \otwo{} A band, and from 19.1 to 9.2 in the strong
\cotwo{} band. Furthermore, even with \cotwo{} fixed, the fitted path redistribution reproduces
each sounding's apparent \cotwo{} deficit (Fig.~\ref{fig:lightpath}f) by shifting 3 to 13\,\% of
detected radiance from the surface-reaching path into elevated path classes. Resolving path length
vertically proves essential: in diagnostic fits to 1{,}500 cloudy spectra, a single path length or
a vertically uniform path-length distribution left 16 to 21\,\% of water-line depth unexplained,
because clouds obscure low-lying water vapor more strongly than well-mixed \cotwo{}.

This same diagnostic clarifies what is and is not recoverable. Even after conditioning on the
\otwo{} A band, the fitted path in the \cotwo{} bands remains uncertain by several percent of the
column; consequently, physics alone cannot constrain \xco{} to 1\ppm{} from an individual cloudy
spectrum. Instead, unlabeled real spectra supply the systematic cross-band structure under real
clouds. Because residual single-sounding cloud errors remain large---and retain a spatially
correlated component across a cloudy scene---we report pool retrievals as 100\,km patch averages
accompanied by an explicit correlated-error variance. Finally, our one-dimensional cloud
simulations do not exhibit this spectral signature (Appendix~\ref{app:simulation}); horizontal
photon transport in heterogeneous cloud decks \citep{massie2021analysis,massie2023insights}, which
one-dimensional radiative transfer omits, offers a plausible explanation.

\subsection{Limits and future directions}
\label{sec:limits}

\paragraph{Pool values are aggregates.}
Our claims beyond the cloud screen apply to patch-level estimates after scene correction and prior
updating. Before the prior update, pool patch means outperform the map on their own strictly
adjacent to accepted soundings (0.88 versus 1.10\ppm{} in error after rescaling). Elsewhere, they
match the map after rescaling while carrying a larger raw RMSE, even though they recover a
regression slope near 0.8 against true anomalies compared with 0.45 for the prior map. Accordingly,
pool values should not be disaggregated, and users assimilating Penumbra with an independent prior
should start from the pre-update estimates provided in the product.

\paragraph{Validation coverage.}
Patch evaluations in Table~\ref{tab:main} require accepted soundings either inside the patch or
within 200\,km along track, whereas 41\,\% of the pool is isolated; likewise, both the scene
correction and the reported uncertainty are trained where accepted soundings occur. Independent
ground truth is also confined to five TCCON stations spanning 34$^\circ$N to 55$^\circ$N, leaving
our tropical and high-latitude assessments reliant on labels derived from operational retrievals.
As a result, the region with the largest coverage gain, north of 60$^\circ$N, remains the least
directly validated.

\paragraph{Global bias offset and averaging kernels.}
With a single constant offset fitted on 2017--2020, the annual mean error across accepted overpasses
remains within 0.13\ppm{} of zero during 2017--2021 for both Penumbra and the operational product,
although individual stations can deviate by up to 1\ppm{} for either method. Because the prior map
inherits the baseline offset of the operational product, which sits 0.2 to 0.3\ppm{} above our
TCCON-anchored estimates, the mean offset of a delivered value shifts slightly with the weight
assigned to the map. Finally, Penumbra does not yet output per-sounding column averaging kernels;
deriving effective vertical sensitivities---for example from perturbations of the clear-sky
Jacobian or from the predicted vertical path weights---is an important next step for assimilation
in atmospheric flux inversions.

\paragraph{Future directions.}
\label{sec:future}
The framework itself is not restricted to the Americas or to OCO-2. Applying Penumbra elsewhere
requires a clear-sky forward model, an instrument line shape, and sounding density sufficient for
neighbors to share \cotwo{}, conditions met by OCO-3 and the planned CO2M constellation
\citep{eldering2019oco,sierk2021copernicus}. Even so, reference transfer has not yet been evaluated
over ocean, and the real-spectrum training terms must be re-fitted on observations from each new
domain. As a next step, testing the pool directly in atmospheric flux inversions that propagate our
reported correlated errors will determine how much these newly recovered regions sharpen surface
flux constraints.

\section{Conclusions}
\label{sec:conclusion}

Penumbra demonstrates that a single physics-constrained neural retrieval can deliver
operational-grade accuracy, substantial gains in cloudy-sky coverage, and two orders of magnitude
higher throughput without training on labeled real spectra. Evaluated on held-out 2021 observations
over the Americas, it achieves parity with the bias-corrected operational product on accepted
soundings at TCCON stations (0.92 versus 0.91\ppm{} per overpass). Beyond the operational quality
filter, delivered estimates for flagged soundings and 100\,km pool aggregates outperform a prior
map across all six 2021 evaluations by 12 to 27\,\%, tripling the usable fraction of land soundings
from 10 to 30\,\%. Because reference transfer restricts exact forward-model runs to roughly one
sounding in ten, retrieval speed increases by a factor of 80 to 116 per sounding relative to
full-physics optimal estimation. These gains stem from separating known clear-sky physics from
empirical cloud behavior: the forward model supplies clear-sky radiative transfer and local \cotwo{}
sensitivity, a fixed light-path layer represents vertical scattering effects, and label-free spatial
constraints allow the network to learn real-cloud behavior directly from observations.

\section*{Data and code availability}
All observational datasets used in this study are publicly available. OCO-2 Level~1B calibrated
science spectra (version~11r, \texttt{OCO2\_L1B\_Science}) \citep{oco2team2019l1b} and Level~2
bias-corrected \xco{} Lite files (version~11.2r, \texttt{OCO2\_L2\_Lite\_FP})
\citep{oco2team2024lite}, together with the accompanying operational preprocessor and meteorological
products, are distributed by the NASA Goddard Earth Sciences Data and Information Services Center
(GES DISC). Ground-based TCCON GGG2020 data for Park Falls, Lamont, Caltech (Pasadena), Edwards,
JPL, East Trout Lake, and Eureka are available from CaltechDATA
\citep{wennberg2022parkfallsr1,wennberg2025lamont,wennberg2022caltech,iraci2022edwards,
wennberg2022jpl,wunch2022easttroutlake,strong2022eureka}. In situ vertical profiles used for
independent evaluation are publicly available from the NOAA Global Monitoring Laboratory for Global
Greenhouse Gas Reference Network aircraft flasks \citep{mckain2022noaa} and AirCore profiles
\citep{baier2021noaa}, from the ORNL Distributed Active Archive Center for ACT-America
\citep{davis2018actamerica}, and from PANGAEA for Amazon vertical profiles \citep{gatti2026co2}.
The Penumbra codebase, trained model weights, and evaluation scripts are currently under
preparation and will be released publicly soon.

\section*{Acknowledgments}
We thank the NASA Orbiting Carbon Observatory-2 (OCO-2) project and science team for producing and
maintaining the OCO-2 observations and operational retrieval products. This work was supported by a
NASA High-End Computing (HEC) grant, which provided computational resources for model training and
evaluation.

\bibliographystyle{plainnat}
\bibliography{references}

\clearpage
\appendix
\section{Simulation details}
\label{app:simulation}

\paragraph{Forward model and drawing rules.}
We generate simulated spectra with a local build of the operational full-physics forward model
(build B11.0.65 using the ABSCO v5.2 absorption tables and static inputs from operational
version~11 processing). Because the standard operational low-streams interpolation (LSI)
configuration \citep{odell2010acceleration} loses accuracy in cloudy scenes, simulated spectra use
LIDORT with an upgraded four-stream low-order model and 16 half-streams at the LSI optical-depth
bins. From each anchor sounding, a simulation inherits the solar and viewing geometry,
meteorological profiles and surface pressure, prior \cotwo{} profile, aerosol types, and instrument
model. Independently of those properties, we draw the \xco{} offset uniformly within $\pm 12$\ppm{}
along with three column-neutral vertical profile-shape modes. In broad draws, clouds are absent,
liquid, ice, or mixed in equal quarters, spanning optical depths from 0.01 to 100; half of the
cloudy draws feature partial cover, computed via the independent pixel approximation by combining a
clear column with a cloudy one. Aerosol optical depth varies within a factor of ten of the
operational prior, surface reflectance is sampled around operational values from neighboring
soundings, and solar-induced chlorophyll fluorescence (SIF) is drawn around the IDP estimate. Small
perturbations to surface pressure and temperature, water-vapor scaling, and wavelength shifts
follow operational retrieval statistics. Horizontal photon transport, stray light, sloping terrain,
and radiation-induced spikes are omitted.

We generate 278{,}164 simulations on 17{,}403 accepted, flagged, and pool anchors from 2017--2019
(16 draws per anchor), training on 245{,}076 and reserving 33{,}088 from 10\,\% of the orbit files
for validation. Replaying 1{,}000 operational retrievals confirms consistency with the operational
build: reduced $\chi^2$ values are 1.22, 1.31, and 1.23 across the three bands versus 1.23, 1.32,
and 1.24 in the official product. By contrast, the clear-sky reference of
Section~\ref{sec:reference-transfer} runs the same forward model in its operational
radiative-transfer mode, including Rayleigh scattering and gas absorption over a gray Lambertian
surface (albedo 0.2) without cloud, aerosol, or fluorescence.

\paragraph{Draws guided by the measured spectrum.}
Among the 16 draws per anchor, four come from the broad distributions above while twelve are guided
by the anchor's observed spectrum. An inverse model proposes an initial scene from eleven spectral
summaries (per-band continuum level and slope, \otwo{} line depths, and cross-band \cotwo{} and
\hto{} ratios). A forward-model emulator then adjusts the cloud, aerosol, and surface state over up
to two steps toward those summaries. Because the \xco{} offset is sampled prior to these
adjustments and held fixed throughout, pre-training checks confirm that its correlation with
quantities read from the real sounding stays below 0.006.

\paragraph{What one-dimensional clouds do not reproduce.}
In unretrieved cloudy soundings, the \cotwo{} bands indicate less air mass than the \otwo{} A band
implies. This discrepancy exceeds the corresponding mismatch on matched simulations and peaks under
bright overcast that appears spatially uniform at the footprint scale. Diagnostic checks spanning
data processing, instrument calibration, forward-model inputs, single-column cloud vertical
structure, and radiative-transfer solver fidelity do not eliminate the discrepancy. Horizontal
photon transport within sub-footprint cloud heterogeneity therefore remains an untested candidate.

\section{Network, training and baselines}
\label{app:training}

\paragraph{Input.}
Because the instrument's eight footprints have slightly different spectral dispersions, we
interpolate $\lnxc$ onto a common wavelength grid. The grid covers the operational fit windows:
0.7592--0.7714\um{} (835 points), 1.5980--1.6178\um{} (652 points), and 2.0477--2.0797\um{} (811
points). A wavelength sample is retained if calibrated radiances do not flag it as bad or
radiation-damaged, it is absent from a per-footprint list of persistent-residual channels, and its
reference transmittance is at least 5\,\%. We fit the per-band quadratic continuum by unweighted
least squares and standardize spectra and continuous context variables using training-simulation
means and standard deviations; invalid samples are set to zero and marked in a second input
channel. Thirty context inputs enter the network: air mass, solar and viewing zenith angles,
surface altitude, meteorological surface pressure and column water vapor, log continuum
signal-to-noise ratio per band, the nine removed continuum coefficients, one-hot indicators for
viewing mode and footprint, and the surface-pressure departure from the meteorological analysis.
Because simulations sample this pressure departure over a wider range (4.3\,hPa) than observed in
reality (about 1\,hPa), we supply the drawn departure perturbed by 1\,hPa of Gaussian noise during
training and set it to zero on real soundings so the network treats the meteorological surface
pressure as accurate to within $\sim$1\,hPa.

\paragraph{Architecture.}
Each band tower takes two channels (value and validity) through five 1-D convolutions
with 32, 64, 64, 96, and 32 channels, kernel sizes 9, 7, 5, 5, and 1, and stride 2 in the second
through fourth layers, each followed by a GELU activation. Flattened tower outputs are projected to
256 units per band and context features to 64 units. The trunk then maps the concatenated 832 units
through a 512-unit layer (with 0.1 dropout) to 256 units. A linear head predicts \xco{}, whereas
the uncertainty head (64 hidden units) reads the stop-gradient trunk output together with the
context. By contrast, the light-path head (128 hidden units) passes gradients into the trunk while
outputting the 26 light-path parameters, with water scale bounded to 0.74--1.35 and path scale to
0.95--1.05. The full network contains 3{,}155{,}170 parameters, including 40{,}090 in the
light-path head.

\paragraph{Light-path layer.}
Among the eight path classes, the first traverses zero absorbing gas (fluorescence, stray light, or
very high-altitude scattering), and the next five backscatter at 21, 42, 63, 79, and 89\,\% of
surface pressure. The seventh reaches the surface directly, while the eighth reaches the surface
after traversing the layer below 63\,\% of surface pressure twice. Layer optical depths in
Eq.~\eqref{eq:layer} are precomputed from forward-model absorption coefficients at 0.01\,cm$^{-1}$
spacing across 19 pressure layers for 64 reference atmospheres covering latitude, season, surface
pressure, and column water. Each sounding selects the nearest reference and rescales it to match
its own surface pressure, column water, and prior \xco{}. For band $b$, the reduced $\chi^2$ misfit
is $\chi^2_b = n_b^{-1}\sum_\lambda \omega_\lambda\,(z-\hat z)^2$ over valid channels with
clear-sky transmittance above 5\,\%, weighted by
$\omega_\lambda^{-1} = (\mathrm{SNR}_b^2\,T^{c}_\lambda)^{-1} + 0.003^2$, where $T^{c}$ denotes
clear-sky transmittance and the constant floor represents model error. The reconstruction loss is
the band average of $10\ln(1+\chi^2_b/10)$.

\paragraph{Training.}
Each optimization step combines 512 simulated spectra (with freshly drawn instrument noise) and
roughly 660 real spectra. Those real spectra comprise accepted soundings from 40 patches, up to six
pool soundings from 40 patches, up to six flagged soundings from 40 patches, 40 cross-orbit patch
pairs (up to four accepted soundings per side), and 63 soundings for the response constraint. In
total, the real training pool holds 45{,}497 accepted, 86{,}253 flagged, and 94{,}687 pool
soundings. Within each step, reconstruction loss is evaluated on 128 simulated and 192 real
spectra; the cross-orbit Huber threshold is 2\ppm. For the response constraint, exact clear-sky
Jacobians exist for 24\,\% of real spectra, while a compact predictor estimates the remaining
Jacobians from the reference spectrum with relative error near 1\,\%. Real-spectrum terms other than
reconstruction ramp up linearly over the first six epochs. Using AdamW
\citep{loshchilov2019decoupled} without early stopping, we train for 72 epochs over the 245{,}076
training simulations (34{,}488 steps) with weight decay $10^{-4}$ and a learning rate
cosine-annealed from $10^{-3}$ to $10^{-5}$. Training one ensemble member takes about 30~minutes on
a single GPU.

\paragraph{Loss terms.}
Trained under a Gaussian error likelihood on simulated spectra, the uncertainty head does not
affect the \xco{} prediction. The neighbor-consistency term computes the pooled within-patch
variance of estimates separately for each class $C$ (accepted, flagged, pool):
\begin{equation}
  \mathcal{L}_{\text{nbr}} \;=\; \sum_{C}\;
  \frac{\sum_{P}\sum_{i\in P_C}\bigl(\hat\delta_i-\bar\delta_{P_C}\bigr)^2}
       {\sum_{P}\bigl(|P_C|-1\bigr)} ,
\end{equation}
where $P_C$ denotes the soundings of class $C$ in patch $P$ and $\bar\delta_{P_C}$ their mean
estimate. Inside Eq.~\eqref{eq:loss}, \xco{} quantities are normalized by 6.93\ppm{} (the standard
deviation of the drawn offset), and loss weights refer to those units. Applied across all three
classes, the response constraint requires a clear-sky perturbation of $d$\ppm{} to shift the output
by $d$\ppm{} because $\vect{h}_i$ represents a clear-sky sensitivity. In cloudy scenes, however, part of
the light does not traverse the full column, which weakens the spectral signature of a true change
and leaves the true-change response slightly below 1.

\paragraph{Ensemble and validation.}
Predictions are averaged across the six ensemble members. The reported standard deviation equals
the square root of mean predicted variance plus ensemble variance, scaled by 1.085 to enclose
68\,\% of validation-simulation errors. Across those 33{,}088 held-out simulations, ensemble RMSE
is 1.14\ppm{} for cloud-free scenes, 1.58\ppm{} for cloud optical depths up to 0.3, 2.84\ppm{}
between 0.3 and 1, and 4.67\ppm{} above 1. For the development-year comparisons in
Section~\ref{sec:res-accepted}, the variants reading 2017 and 2018 follow the same training
procedure with the same simulations and real spectra from the other two years of 2017--2019.

\paragraph{\cotwo{} response test.}
For soundings with an exact clear-sky Jacobian $\vect{h}$, we evaluate response via the central
difference $[f_\theta(z+d\vect{h})-f_\theta(z-d\vect{h})]/2d$ with $d=1$\ppm. Because this mirrors
the perturbation used during training, the test confirms out-of-sample adherence to the constraint
for Penumbra while measuring an unconstrained sensitivity for the baselines.
Table~\ref{tab:ablation} reports medians over 124{,}647 development soundings across all three
classes. On real 2019 spectra held out from training, the adopted network maintains a mean response
of 0.97 on accepted soundings and 0.97 on the pool.

\paragraph{Transferred reference.}
Exact references are placed on a fixed along-track grid: within each 32-frame window of a footprint
that holds an accepted, flagged, or pool sounding, the sounding nearest the window center receives
an exact run. Remaining soundings interpolate between the nearest preceding and following grid
references on the same footprint (up to 64 frames away) using inverse-distance weights. Each
borrowed reference is adjusted by averaging two predictions of the change in $\ln c$ between
soundings. The first is a first-order transfer using predicted Jacobians for differences in prior
\xco{}, water, surface pressure, and temperature (refined by a dual-state network), and the second
is the difference between reference emulators evaluated at the two states using location and day of
year to approximate the clear-sky forward-model difference (location and date never enter the
retrieval network). Whenever the distance-weighted meteorological surface-pressure difference from
the two bracketing references exceeds 50\,hPa, the sounding receives its own exact run. Across
accepted, flagged, and pool soundings of 2017--2020, 6.7\,\% undergo exact runs due to the grid and
another 4.4\,\% due to the pressure threshold. In 2021, reference transfer shifts estimates with a
standard deviation of 0.03\ppm{} next to a grid node and 0.13\ppm{} at maximum separation, whereas
soundings routed to exact runs by the pressure rule would otherwise have shifted by 0.21\ppm. In
the Section~\ref{sec:res-speed} benchmark, an exact run averages 16.2\,s, reference transfer takes
3\,ms per sounding on a single CPU core, and six-member ensemble inference takes 6\,ms on CPU (or
0.02\,ms on a GPU, as timed in the benchmark).

\paragraph{Post-processing details.}
Post-processing tree models are histogram-based gradient-boosted regressors with 120 to 300 trees.
The scene-correction model uses 64 predictors: the two band ratios, the cloud-screen
surface-pressure difference and flag, predicted network uncertainty, four light-path outputs,
twelve scene variables (including air mass, column water vapor, neighboring retrieval albedo, and
nearby cloud-flagged fraction), forty diagnostics from one linearized retrieval step with
network-predicted Jacobians (requiring zero forward runs), and three patch descriptors. Its target
is the sounding's estimate minus the mean estimate of accepted soundings in the same patch. For the
91\,\% of fit soundings with accepted neighbors, this target involves solely Penumbra estimates;
for the remaining 9\,\%, operational values enter via the along-track label shifted to our accepted
level. Where a patch lacks accepted soundings, the along-track label likewise serves as reference
for the patch-level correction and as target for the correlated-variance model. Fitted on
527{,}929 flagged and pool soundings from 2017--2020 using out-of-year network variants, this
correction leaves accepted soundings unmodified. A second tree model estimates the variance of each
corrected sounding. After computing the inverse-variance average of a patch's pool soundings, a
patch-level tree model removes remaining predictable bias in the mean. Correlated patch error
variance is predicted from patch-level features, scaled to match the correlated error observed at
TCCON in 2017--2020 (1.39\ppm$^2$), and floored at half that value. For accepted soundings, the
predicted standard deviation is scaled by 1.10 to standardize 2017--2020 TCCON errors.

The prior map takes latitude, longitude, month as an annual harmonic, and a region index, plus a
linear trend of $-0.127$\ppm{} per year to capture the drift of operational values relative to the
prior \xco{}. Fitted to labels from 32{,}991 patches, it assigns out-of-fold residual standard
deviations by region and season ranging from 0.86 to 1.54\ppm. Under the prior update, posterior
variance becomes $(1/\sigma^2+1/\sigma_\mu^2)^{-1}$. Each delivered estimate responds to the
measurement with sensitivity $\sigma_\mu^2/(\sigma^2+\sigma_\mu^2)$, which we report alongside
both the pre-update and delivered values.

\paragraph{Baselines.}
For single-sounding comparisons and equal-footing prior updates, each method receives one global
offset and one uncertainty scale factor fitted on 2017--2020 development overpasses ($+0.14$\ppm{}
and 1.10 for Penumbra; $-0.16$\ppm{} and 2.49 for the operational product). Penumbra's offset here
differs marginally from the $+0.15$\ppm{} used in the main pipeline because it is fitted to
unweighted overpass means. Matching our architecture and inputs apart from the simulation-only
surface-pressure departure, the \emph{label-supervised} baselines comprise two three-member
ensembles. The first trains on 112{,}701 accepted soundings of 2017--2020 labeled by bias-corrected
operational \xco{}, excluding station soundings and their patches. The second starts from the
first, freezes the band towers and context embedding, and fine-tunes the upper layers on 21{,}088
pairs of accepted soundings and kernel-adjusted TCCON values across 723 development overpasses of
2017--2020. Relative to the published systems they represent, these baselines train on fewer
soundings and, in the case of \citet{keely2025conditional}, on real rather than simulated spectra.

Following \citet{meng2022learning} in simplified form, the \emph{per-sounding Bayesian fit}
optimizes a 39-parameter state through a surrogate forward operator trained on simulations with
cloud optical depth at most 0.3. Rather than using a learned generative prior with sampling, it
places a Gaussian prior on \xco{} from the prior map and takes the posterior mode across five
starting points with a Laplace covariance approximation. Ablations follow the adopted training
setup with individual loss terms omitted. In the cloud transplant, half of the cloud-free simulated
spectra receive the measured spectral difference between a 2017--2018 pool or flagged sounding and
accepted soundings in its patch while keeping the simulated \xco{} label. For the \emph{forced
operational retrieval}, we executed the standard operational code without cloud screening across
the 27{,}699 pool soundings in 2021 station overpasses, of which 16{,}619 converged. Taking
pre-bias-correction \xco{} (applying the operational bias correction degrades error after rescaling
further to 1.54\ppm), discarding values more than 20\ppm{} from the prior, and retaining overpasses
with at least three converged soundings, we score errors after rescaling so constant bias does not
penalize the method.

\section{Evaluation data and protocol}
\label{app:tests}

\paragraph{Sounding classes and cloud screens.}
Table~\ref{tab:data} summarizes the sounding classes of Section~\ref{sec:classes} and our data
splits. Assuming a clear sky, the A-Band Preprocessor (ABP) fits the \otwo{} A band and compares
retrieved surface pressure with the meteorological estimate; because clouds shorten the photon
path, an absolute difference $|\Delta P_{\text{ABP}}| > 25$\,hPa flags a sounding as cloudy. The
IMAP-DOAS Preprocessor (IDP) retrieves \cotwo{} and \hto{} columns separately from the weak and
strong \cotwo{} bands under clear-sky assumptions; scattering-induced path differences push the
band ratios away from 1 in the presence of cloud or aerosol \citep{taylor2016orbiting}. Among pool soundings,
6.70~million pass the A-band screen while 4.02~million fail with $|\Delta P_{\text{ABP}}| < 50$\,hPa,
and a subset may represent non-converged operational retrievals. The 2017--2020 archive totals
86.6~million soundings across 287{,}043 patches (median 349 soundings per patch), and the 2021
archive contains 21.2~million soundings. Although 13.9\,\% of the pool consists of nadir footprints
straddling a coastline, our training and evaluation sets retain pure land footprints exclusively.

\begin{table}[!htbp]
\centering
\caption{Sounding classes and data sets. Shares refer to the 86.6~million soundings over land in
the Americas in 2017--2020. A patch is a 100\,km along-track segment of one orbit. The evaluation sets of 2021 hold up to 100 accepted and 40 flagged soundings per overpass (up to 200 accepted where the boxes of Edwards and Pasadena overlap), up to 8 accepted, 12
flagged and 40 pool soundings per patch, and all pool soundings of an overpass.}
\label{tab:data}
\small
\setlength{\tabcolsep}{4pt}
\begin{tabularx}{\textwidth}{@{}l Y r r Y@{}}
\toprule
\multicolumn{5}{@{}l}{\textbf{A. Sounding classes}}\\
\midrule
Class & Definition & Share & Soundings & Penumbra delivers \\
\midrule
Accepted & operational value, good quality flag (QF = 0) & 10.3\,\% & 8.92\,M & one value per sounding \\
Flagged & operational value, bad quality flag (QF = 1) & 7.7\,\% & 6.68\,M & one value per sounding \\
Pool & unretrieved; fixed preprocessor bounds (Section~\ref{sec:classes}) & 12.4\,\% & 10.72\,M & one value per patch with $\geq 10$ pool soundings \\
Outside the pool & unretrieved; outside preprocessor bounds (too optically thick) & 69.6\,\% & 60.28\,M & -- \\
\midrule
\multicolumn{5}{@{}l}{\textbf{B. Data sets}}\\
\midrule
Data set & Content & Years & Soundings & Truth and role \\
\midrule
Simulations & forward-model spectra on real anchor soundings & 2017--19 & 278{,}164 & drawn \xco{};
supervised training (33{,}088 held out) \\
Real spectra & accepted, flagged and pool soundings in patches & 2017--18 & 226{,}437 & none;
label-free training \\
Development sets & all real soundings with a clear-sky reference (patches, station overpasses, aircraft coincidences; includes the row above) & 2017--20 & 0.88\,M & patch labels, TCCON;
post-processing fits and design choices \\
Station overpasses & all overpasses at five TCCON stations & 2021 & 50{,}825 & TCCON; held-out test (232 overpasses, 226 scored) \\
Patches, $\geq$5 accepted & stratified by region and season & 2021 & 105{,}082 & patch label;
held-out test (2{,}000 patches) \\
Patches, $<$5 accepted & stratified by region and season & 2021 & 75{,}963 & along-track label;
held-out test (2{,}000 patches) \\
Speed test & two stretches of orbit & 2021 & 2{,}014 & timing only \\
\bottomrule
\end{tabularx}
\end{table}

\paragraph{Ground stations and labels.}
Seven TCCON stations operated within our study domain during 2017--2021: Park Falls, Lamont,
Caltech (Pasadena), Edwards, JPL, East Trout Lake, and Eureka
\citep{wennberg2022parkfallsr1,wennberg2025lamont,wennberg2022caltech,iraci2022edwards,
wennberg2022jpl,wunch2022easttroutlake,strong2022eureka}. Coincidence boxes span $2.5^\circ$ in
latitude by $5^\circ$ in longitude for most stations, following \citet{das2025comparisons}, and
$1^\circ \times 1^\circ$ at Pasadena and $2^\circ \times 10^\circ$ at Edwards (twice the full
widths used by \citet{das2025comparisons} at those two sites). For each sounding, we adjust the
mean of TCCON spectra collected within one hour to the sounding's prior and column averaging kernel
\citep{wunch2011method}, taking the kernel from the operational retrieval. Where a sounding lacks
its own kernel, we substitute the mean kernel of accepted soundings in the overpass or, if
unavailable, that of other retrieved soundings in the overpass.

\paragraph{Splits and evaluation protocol.}
Because consistency losses allow a network to memorize its training patches (reducing within-patch
scatter of accepted estimates from 0.93\ppm{} on unseen patches to 0.36\ppm{} on training patches),
downstream models fitted to a given year's predictions use a network variant trained without that
year. Thus, 2017 is predicted by a variant trained on 2018--2019 real spectra, 2018 by a variant
trained on 2017 and 2019, and 2019--2021 by the adopted network. All 2021 observations were held
out until the network, post-processing pipeline, baselines, and evaluation sampling rules were
finalized. Evaluation sets for 2021 were selected purely from sounding identifiers and cloud-screen
outputs: all 2021 TCCON overpasses, 2{,}000 patches with at least five accepted soundings and
2{,}000 with fewer (both stratified by region and season), plus two orbit segments (2{,}014
soundings) for timing. Block-bootstrapping entire days instead of overpasses or patches widens
patch confidence intervals by up to 17\,\% without altering any conclusions.

\section{Additional results}
\label{app:tables}

\paragraph{Evaluation sets and reported uncertainty.}
Table~\ref{tab:unc} lists sample counts behind each row of Table~\ref{tab:main} alongside the
fraction of truths falling within one and two reported standard deviations. In addition, the
development rows of Table~\ref{tab:main}A comprise 624 overpasses, while the forced operational
comparison spans 199.

\begin{table}[ht]
\centering
\caption{Evaluation sets of 2021, with the number of overpasses (ground stations) or patches and
the coverage of the reported uncertainty of the delivered values (nominal 68 and 95\,\%).}
\label{tab:unc}
\footnotesize
\begin{tabularx}{\textwidth}{@{}l Y r c c@{}}
\toprule
Soundings & Truth & Number & Within 1 s.d. & Within 2 s.d. \\
\midrule
Accepted & ground stations (TCCON) & 181 & 76\,\% & 96\,\% \\
Flagged & ground stations (TCCON) & 207 & 70\,\% & 91\,\% \\
Flagged & accepted soundings in the same patch & 1{,}981 & 68\,\% & 92\,\% \\
Pool & ground stations (TCCON) & 186 & 69\,\% & 90\,\% \\
Pool & accepted soundings in the same patch & 1{,}870 & 68\,\% & 93\,\% \\
Pool & accepted soundings 50--200\,km along the orbit & 1{,}792 & 62\,\% & 89\,\% \\
 & \quad the same, no accepted sounding in the patch & 1{,}105 & -- & -- \\
\bottomrule
\end{tabularx}
\end{table}

\paragraph{Accepted soundings by year.}
Table~\ref{tab:peryear} breaks down the operational comparison by individual year. Development-year
gains are strongest in 2018 and 2019, marginal in 2020, and absent in 2017, retaining the same sign
under whole station-year resampling (\ci{-0.13}{-0.03}). At the single-sounding level in 2021, RMSE
is 1.20\ppm{} for Penumbra, 1.21\ppm{} for the operational product, and 1.81\ppm{} for the
operational retrieval prior to bias correction. Across 2017--2021, annual mean errors over accepted
overpasses stay within $-0.07$ to $+0.13$\ppm{} for Penumbra and $-0.09$ to $+0.13$\ppm{} for the
operational product, although single stations exhibit larger offsets; at Pasadena (within the Los
Angeles urban basin) in 2021, both products sit roughly 1\ppm{} low.

\begin{table}[ht]
\centering
\caption{Accepted soundings at five TCCON stations by year (East Trout Lake, Edwards, Lamont, Park
Falls, Pasadena), with the number $n$ of overpasses. Left, error after rescaling of overpass means
of the single-sounding estimates. Right, RMSE after the same prior update, where for 2017--2020
the prior map and all constants are fitted without the scored year. Differences are Penumbra minus
operational with 95\,\% intervals. Only 2021 is held out from development.}
\label{tab:peryear}
\footnotesize
\setlength{\tabcolsep}{0pt}
\begin{tabular*}{\textwidth}{@{\extracolsep{\fill}}l r c c l c c l@{}}
\toprule
 & & \multicolumn{3}{c}{Single-sounding estimates} & \multicolumn{3}{c}{After the same prior update} \\
\cmidrule(lr){3-5}\cmidrule(lr){6-8}
Year & $n$ & Penumbra & Operational & Difference & Penumbra & Operational & Difference \\
\midrule
2017 & 107 & 0.842 & 0.817 & $+0.025$ \ci{-0.055}{+0.098} & 0.783 & 0.782 & $+0.001$ \ci{-0.051}{+0.047} \\
2018 & 158 & 0.786 & 0.905 & $-0.119$ \ci{-0.204}{-0.028} & 0.690 & 0.746 & $-0.056$ \ci{-0.114}{+0.005} \\
2019 & 160 & 0.668 & 0.812 & $-0.144$ \ci{-0.209}{-0.081} & 0.637 & 0.730 & $-0.093$ \ci{-0.145}{-0.046} \\
2020 & 199 & 0.810 & 0.871 & $-0.061$ \ci{-0.123}{+0.007} & 0.807 & 0.852 & $-0.045$ \ci{-0.095}{+0.004} \\
\midrule
2017--2020 & 624 & 0.778 & 0.865 & $-0.087$ \ci{-0.126}{-0.047} & 0.733 & 0.784 & $-0.050$ \ci{-0.077}{-0.023} \\
\midrule
2021 (sealed) & 181 & 0.916 & 0.909 & $+0.007$ \ci{-0.060}{+0.066} & 0.827 & 0.851 & $-0.024$ \ci{-0.061}{+0.013} \\
\bottomrule
\end{tabular*}
\end{table}

\paragraph{Pool by region, season and distance.}
Table~\ref{tab:region} disaggregates the two pool populations from Table~\ref{tab:main}B by region
and season. In the withheld-accepted experiment of Section~\ref{sec:res-coverage}, incorporating
local pool values reduces prediction RMSE by 0.09, 0.13, and 0.19\ppm{} across withheld segments of
20, 40, and 60\,km, and by 0.07 to 0.20\ppm{} for full patches separated by 10 to 800\,km from the
nearest accepted soundings, with each confidence interval excluding zero.

\begin{table}[ht]
\centering
\caption{Delivered pool values in 2021 by region and season, as RMSE (ppm) of Penumbra and of the
prior map with the number $n$ of patches.}
\label{tab:region}
\footnotesize
\setlength{\tabcolsep}{0pt}
\begin{tabular*}{\textwidth}{@{\extracolsep{\fill}}l r c c r c c@{}}
\toprule
 & \multicolumn{3}{c}{Truth in the same patch} & \multicolumn{3}{c}{Truth along the orbit} \\
\cmidrule(lr){2-4}\cmidrule(lr){5-7}
 & $n$ & Penumbra & Prior map & $n$ & Penumbra & Prior map \\
\midrule
North America & 447 & 1.035 & 1.399 & 446 & 1.336 & 1.528 \\
Central America and Caribbean & 457 & 0.723 & 0.919 & 443 & 0.902 & 1.005 \\
Amazon & 431 & 0.886 & 1.126 & 419 & 0.946 & 1.072 \\
Rest of South America & 535 & 0.672 & 0.926 & 484 & 0.914 & 1.064 \\
\midrule
December--February & 421 & 0.890 & 0.979 & 420 & 1.045 & 1.083 \\
March--May & 483 & 0.699 & 0.958 & 463 & 0.930 & 1.037 \\
June--August & 486 & 0.834 & 1.251 & 443 & 1.019 & 1.272 \\
September--November & 480 & 0.901 & 1.172 & 466 & 1.149 & 1.320 \\
\bottomrule
\end{tabular*}
\end{table}

\paragraph{Aircraft profiles.}
As an independent check outside both satellite and TCCON records, we converted aircraft and AirCore
profiles \citep{sweeney2015seasonal,davis2021atmospheric,karion2010aircore,gatti2021amazonia,
mckain2022noaa,baier2021noaa,davis2018actamerica,gatti2026co2} into column averages using the prior
and averaging kernel of each coincident sounding. During 2021, 24 profiles coincide with
overpasses: across 7 accepted overpasses, RMSE is 1.38\ppm{} for Penumbra versus 1.51\ppm{} for the
operational product. Across 10 flagged and 13 pool overpasses, raw single-sounding estimates prior
to correction yield 1.35 and 1.34\ppm, compared with 0.85 and 0.98\ppm{} for the prior map. While
the flagged difference is statistically significant ($+0.50$\ppm{} \ci{+0.12}{+0.81}) and the pool
difference is not, both results align with the Section~\ref{sec:limits} caveats for uncorrected,
pre-update values; given the small sample size, stronger inferences are unwarranted.

\paragraph{The published validation statistic.}
\label{app:das}
Whereas our tables report operational TCCON errors between 0.9 and 1\ppm, published validation
cites 0.85\ppm{} \citep{das2025comparisons}. To confirm that both figures reflect the same
underlying product, we reproduced the published protocol across our seven stations for 2017--2020.
For each overpass, we took the median difference between accepted soundings in the box and
kernel-adjusted TCCON, keeping overpasses with at least 100 accepted soundings and at least 15
TCCON spectra within one hour. Over 353 qualifying overpasses, this rule yields a mean difference
of $+0.01$\ppm{} and a standard deviation of 0.82\ppm, matching the published
$-0.03 \pm 0.85$\ppm{}. The scatter depends strongly on the minimum sounding threshold: admitting
overpasses with as few as five accepted soundings, as in our main tables, raises the standard
deviation to 1.04\ppm, whereas requiring 100 lowers it to 0.82\ppm. The two statistics therefore
summarize the same product on different overpass subsets.

\end{document}